\pdfoutput=1
 
\documentclass[letterpaper, 10 pt, conference]{ieeeconf}  

\makeatletter
\newcommand*\titleheader[1]{\gdef\@titleheader{#1}}
\AtBeginDocument{%
  \let\st@red@title\@title
  \def\@title{%
    \bgroup\normalfont\scriptsize\@titleheader\par\egroup
    \vskip1.5em\st@red@title}
}
\makeatother
\titleheader{This work has been submitted to the IEEE for possible publication. Copyright may be transferred without notice, after which this version may no longer be accessible.}

\IEEEoverridecommandlockouts                              

\usepackage{epsfig} 
\usepackage{times} 
\usepackage{amsmath} 
\usepackage{amssymb}  
\usepackage{graphicx}
\usepackage{hyperref}
\usepackage{xcolor}
\usepackage{multirow}
\usepackage{siunitx}
\usepackage{array}
\usepackage{diagbox}
\usepackage{adjustbox}
\usepackage{booktabs}
\usepackage{todonotes}
\usepackage{pifont}
\usepackage{amssymb}
 \usepackage{url}
\newcommand{\xmark}{\ding{55}}%
\usepackage{colortbl}

\usepackage{amsmath}
\DeclareMathOperator*{\argmax}{argmax} 

\newcommand{\etal}{\emph{et~al.~}}

\newcolumntype{P}[1]{>{\centering\arraybackslash}p{#1}}

\definecolor{road}{RGB}{200,100,50}
\definecolor{sidewalk}{RGB}{255,220,200}
\definecolor{curb}{RGB}{200,49,254}
\definecolor{marking}{RGB}{230,120,10}
\definecolor{terrain}{RGB}{255,0,0}

\renewcommand{\baselinestretch}{0.985}

\newcolumntype{R}[2]{%
    >{\adjustbox{angle=#1,lap=\width-(#2)}\bgroup}%
    l%
    <{\egroup}%
}

\title{\LARGE \bf
Lane Graph Estimation \\ for Scene Understanding in Urban Driving}

\author{Jannik Z\"urn$^{*}$, Johan Vertens$^{*}$, and Wolfram Burgard
\thanks{$^{*}$These authors contributed equally. All authors are with the University of Freiburg, Germany. Wolfram
Burgard is also with the Toyota Research Institute, Los Altos, USA. Corresponding author: {\tt\small zuern@informatik.uni-freiburg.de}}%
}

 \usepackage{fancyhdr}

\begin{document}

\maketitle

\thispagestyle{empty}
\pagestyle{empty}

\begin{abstract}
    Lane-level scene annotations provide invaluable data in autonomous vehicles for trajectory planning in complex environments such as urban areas and cities. However, obtaining such data is time-consuming and expensive since lane annotations have to be annotated manually by humans and are as such hard to scale to large areas. In this work, we propose a novel approach for lane geometry estimation from bird's-eye-view images. We formulate the problem of lane shape and lane connections estimation as a graph estimation problem where lane anchor points are graph nodes and lane segments are graph edges. We train a graph estimation model on multimodal bird's-eye-view data processed from the popular NuScenes dataset and its map expansion pack. We furthermore estimate the direction of the lane connection for each lane segment with a separate model which results in a directed lane graph. We illustrate the performance of our LaneGraphNet model on the challenging NuScenes dataset and provide extensive qualitative and quantitative evaluation. Our model shows promising performance for most evaluated urban scenes and can serve as a step towards automated generation of HD lane annotations for autonomous driving.
   \end{abstract}

\section{Introduction}
\label{sec:introduction}

HD maps play an important role for autonomous navigation~\cite{pannen2020keep, kim2021hd}, in particular in semi-structured and dynamic environments such as urban areas. In this work, we are interested in learning to infer geometric annotations for lane centerlines which are integral to HD maps and which allow autonomous vehicles to perform planning tasks. Being able to access the exact location of lane anchor points and lane directions enables autonomous vehicles to determine legal trajectories from the current position to goal points, both in map-based autonomous driving and in mapless autonomous driving. Despite their utility for autonomous driving, lane-level annotations are expensive and time-consuming to obtain and to maintain. Furthermore, they can be outdated due to construction works or other changes to the road and its surroundings~\cite{pannen2020keep}. Due to the time-consuming manual annotation process, obtaining large-scale lane-level annotations remains a challenge to this day and poses a bottleneck for widespread deployment of fully autonomous driving. Therefore, automatic inference of lane centerline annotations from vehicle sensor readings is an important step towards achieving autonomous driving at scale.

In recent years, many works considered automatic or semi-automatic annotation of road data \cite{bastani2018roadtracer, batra2019improved, mattyus2017deeproadmapper, tan2020vecroad}. Some works focused on extracting road line segments from overhead imagery \cite{batra2019improved,he2020sat2graph,tan2020vecroad,van2018spacenet}, with the disadvantage that lane-level annotations cannot be resolved. Such approaches, however, are not sufficient for autonomous driving and planning as the road network alone does not provide enough data for the vehicle to perform fine-grained planning. Considering the complex topology that street-networks can form, it is necessary to resolve the lane-level connectivity within a given street or road. Other works focused on extracting the exact geometry of the road boundaries \cite{beck2014non}. Few works considered the more complex problem of extracting the precise location of lanes and lane boundaries from ego-vehicle sensor data \cite{beck2014non, homayounfar2019dagmapper, liang2019convolutional}. Extracting precise lane-level geometric annotations for complex urban scenes, including lane direction estimation, remains a challenge.

\begin{figure}
\centering
\includegraphics[width=\linewidth]{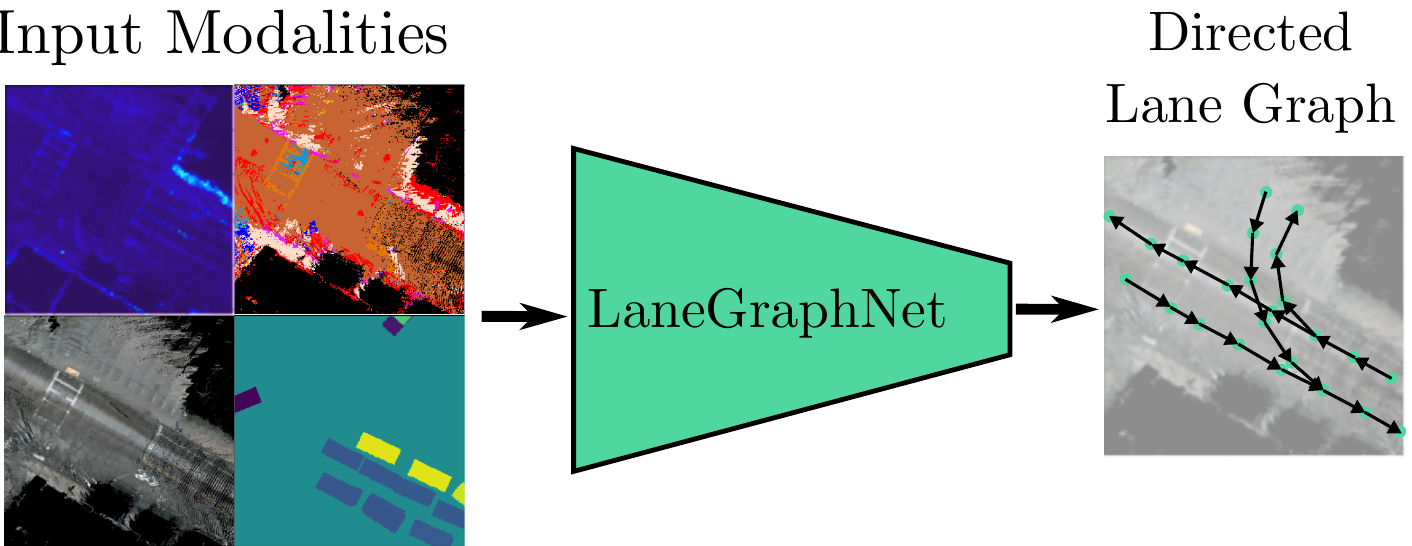}
\caption{Our LaneGraphNet model extracts lane anchor points and unilateral connections between the lane anchor points from multimodal birds-eye-view images in challenging urban environments.}
\label{fig:covergirl}
\end{figure}

In this work, we take a step towards automated lane-level annotations by inferring the underlying graph structure of lanes from multimodal bird's-eye-view image data in an end-to-end fashion. Inspired by recent works on scene graph estimation, we adapt a Graph R-CNN \cite{yang2018graph} model as the backbone for our LaneGraphNet framework. We use our framework to estimate the position and orientation of lanes for a given scene and exclusively leverage input data obtained from a vehicle and do not rely on overhead image data. Our model is trained on a training area with supervision from a HD map and is tested on a different area which serves as the testing set. The inputs of our LaneGraphNet model comprises multiple modalities, including LiDAR intensities, RGB, semantics, and vehicles. While the LiDAR intensities and the RGB pixel values contain raw sensor data, the semantic input layer is the output of a semantic segmentation model trained on a separate dataset and used for inference on the RGB images from the NuScenes dataset. These semantic pixels are projected onto the LiDAR point cloud and used as an additional input modality. Finally, we use a pre-trained 3D object detection model on the LiDAR point cloud to estimate the positions of vehicles in the scenes and project them into the BEV representation as well.

We conduct extensive experiments and demonstrate that our approach is capable of providing highly accurate lane graphs for most evaluated scenes. We also argue that even inaccurate predictions can serve as an initial guess for a human annotator, resulting in a much smaller annotation time compared to creating the graph from scratch. We demonstrate the effectiveness of our approach on the challenging NuScenes dataset. Our approach shows promising results for lane graph estimation in highly complex urban scenes, taking a step towards automatic lane annotation. In summary, the contributions of this work are:

\begin{itemize}
    \item A novel graph-based approach for high-level lane topography estimation.
    \item A novel adaptive graph parameterization technique suitable adapting the ground truth graph node positions to the predictions of the model.
    \item A novel lane graph estimation dataset derived from the highly popular NuScenes dataset, named \textit{NuScenesGraph}, potentially serving as a future benchmark for subsequent works.
    \item Extensive evaluation and ablation studies of our approach.
\end{itemize}

\section{Related Works}
\label{sec:relatedworks}

\subsection{Road Network Estimation}

In the past years, road- and lane network estimation began to attract more attention in the computer vision community. Hereby, most works focus on leveraging aerial RGB images to estimate the structure of road networks~\cite{marmanis2016semantic, mnih2010learning}. With the advent of Deep Learning, many approaches began to consider deep neural architectures for image segmentation and graph estimation~\cite{paz2015variational, bastani2018roadtracer,  chu2019neural, mattyus2017deeproadmapper, bai2018deep}. Mattyus \etal \cite{mattyus2017deeproadmapper} propose to train a semantic segmentation network on aerial images with binary segmentation. They post-process the skeletonized softmax of the network prediction and complete the road graph based on road proposals from the A$^*$ planning algorithm. Similarly, Bastani \etal \cite{bastani2018roadtracer} propose to post-process the output of a binary segmentation model through morphological thinning and the Douglas-Peucker method. Based on simulated data, Meyer \etal \cite{meyer2019anytime} propose a lane-level intersection estimator, leveraging birds-eve-view tracking of traffic participants for intersection graph estimation. Li \etal \cite{li2019topological} use an RNN to build building polygons and road graphs from overhead images, while Chu \etal \cite{chu2019neural} propose to iteratively build a road graph from overhead images. Convolutional recurrent network for road boundary extraction were investigated by Liang \etal \cite{liang2019convolutional} where they jointly use CNNs and RNNs to extract road boundary polylines from LiDAR and RGB image data recorded with an on-board camera and projected into a BEV representation.

\subsection{Lane Network Estimation}

Few recent works considered reasoning about the network structure for roads on a lane-level view \cite{homayounfar2018hierarchical, homayounfar2019dagmapper, sela20203d, liang2020learning}. Homayounfar \etal \cite{homayounfar2018hierarchical} leverage hierarchical recurrent attention networks to build a disjoint graph of lane boundaries from projected LiDAR point clouds for stretches of road without intersections.
Perhaps the most similar work to ours is DAGMapper \cite{homayounfar2019dagmapper} by Homayounfar \etal The authors propose a network architecture that sequentially processes an aggregated LiDAR intensity image by following lane boundaries, thereby producing a directed acyclic graph parameterizing the connectivity of the lanes. In our work, in contrast, each graph node can be estimated separately, without having to necessarily place a preceding node correctly. Furthermore, we evaluate our approach in a more challenging urban environment, in contrast to highly structured highways. More recently, Liang \etal \cite{liang2020learning} propose a vehicle motion forecasting approach leveraging Graph-Convolutional Neural Networks to learn node representations for each graph node that help predicting vehicle movements into the future. They use the ground truth graph node positions, however, and do not predict their position as part of the model inference process.

Compared to the aforementioned approaches, we leverage BEV image data both directly from raw sensor data (RGB and LiDAR) and processed inputs from pre-trained segmentation and detection models to provide additional input for our approach. We furthermore explicitly reason about the direction of each graph edge from context and therefore do not require a sequential model  to represent the graph direction from the anchor point output order. Finally, we apply our model on a challenging dataset with complex urban scenes compared to highly structured highway scenes.

\section{Technical Approach}
\label{sec:approach}

We interpret the problem of lane graph estimation as a directed graph estimation problem. We aim at obtaining a directed lane graph $G = G(v_i,e_i)$ for a given birds-eye-view (BEV) scene, consisting of the graph vertices $v_i$ that correspond to lane anchor points and the directed graph edges $e_i$ correspond to connections between lane anchor points that are directly accessible from each another without breaking traffic rules. The input to our model is a set of pixel-aligned BEV image modalities obtained with the vehicle sensors. We evaluate our model with the modalities LiDAR intensities, RGB color, semantic classes, and vehicle orientations. In the following, we detail each component of our approach.

\begin{figure*}
\centering
\includegraphics[width=\linewidth]{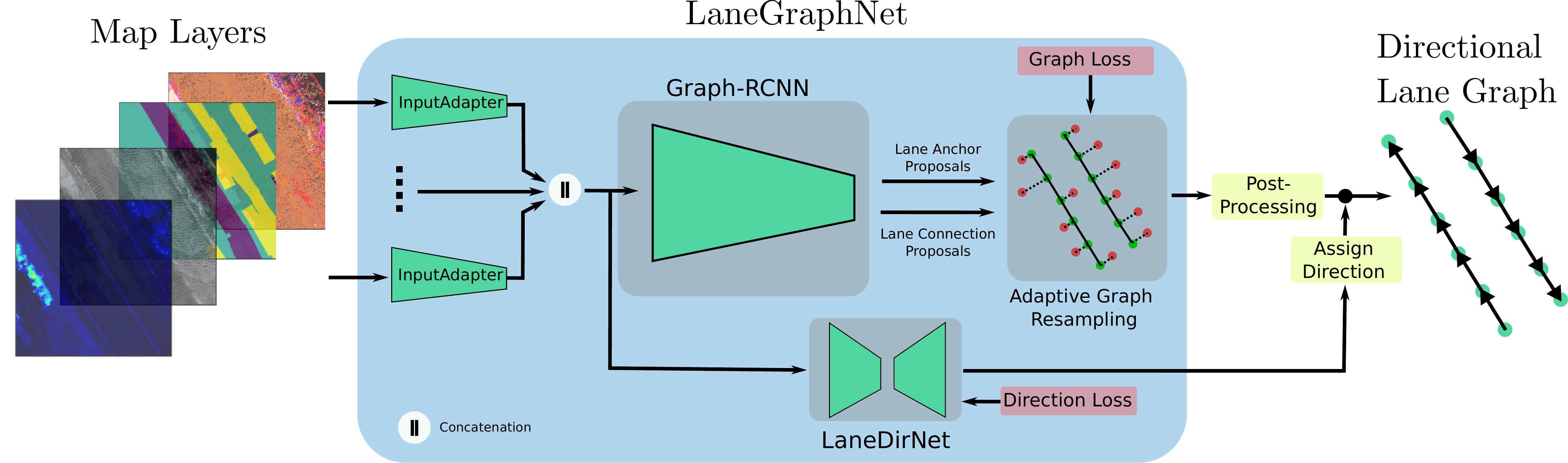}
\caption{Our LaneGraphNet takes LiDAR, RGB, vehicles, and semantics as input. We first extract relevant features from the input modalities using one feature extractor for each modality, we then concatenate the features and use them as inputs for a Graph-RCNN backbone which predicts lane anchors and lane segment relationships. The model uses a supervised loss obtained from the dynamically adapted GT graph. We also feed the concatenated features into our LaneDirNet model which predicts for each pixel in the input map layer the direction}
\label{fig:approach}
\end{figure*}

\subsection{Dataset Generation}

To provide supervision for our LaneGraphNet, we utilize the publicly available NuScenes~\cite{caesar2020nuscenes} dataset. For our experiments, we use all scenes from the Boston-Seaport map. For each scene, we parse the provided lane annotations from the NuScenes map data into a geo-referenced map coordinate frame. The lane annotations entail the lane anchor points and all existing connections between anchor points. Each lane is thus defined by a set of anchor points, which we interpret as graph nodes, and line segments connecting pairs of points, which we interpret as graph edges. We re-sample the lane anchor points such that the maximum distance of one graph node to the next connected node is $\SI{2}{m}$. This representation serves as the graph ground truth graph for all experiments. In the following, we detail the four BEV image modalities used as the input to our model.

The LiDAR modality is obtained by accumulating the LiDAR point clouds over all available time steps in the scene and projecting them into the BEV representation. To remove any moving objects from the accumulated point clouds, we discretize the ground plane into grid cells and for each ground grid cell, we only keep the point with the lowest $z$-coordinate value and discard all other points. This heuristic robustly removes all moving objects in a given scene while ignoring static scene components. We further project the RGB camera images onto the LiDAR point clouds and thus obtain additional RGB pixel values for all accumulated LiDAR points. We furthermore infer semantic classes for each camera image pixel from a semantic segmentation model, pre-trained on the Mapillary dataset. We use a color-coding to convert the class-labels into discriminative class-specific RGB colors. We use the same Camera-to-LiDAR projection pipeline as with the RGB pixels. Finally, we produce an additional BEV modality by using the pre-trained 3-D object detection model CenterPoint \cite{yin2020center} on the LiDAR point clouds from the scene. To this end, we accumulate 10 consecutive LiDAR scans into one point cloud and infer all 3-D object bounding boxes for that accumulated scan. We remove all non-vehicle detections and project the remaining bounding boxes into a BEV representation and rasterize the result to an image. All pixels without a vehicle bounding box are labelled -1, while pixels with a bounding box are assigned a value according to the rotation angle of the bounding box w.r.t. the global map north. This angle is normalized to be in the interval $[0, 1)$. For all modalities we chose an image resolution of 20 cm per pixel. Tab. \ref{tab:dataset-details} lists the key data for our derived dataset \textit{NuScenesGraph}.

\begin{figure}
\centering
\includegraphics[width=\linewidth]{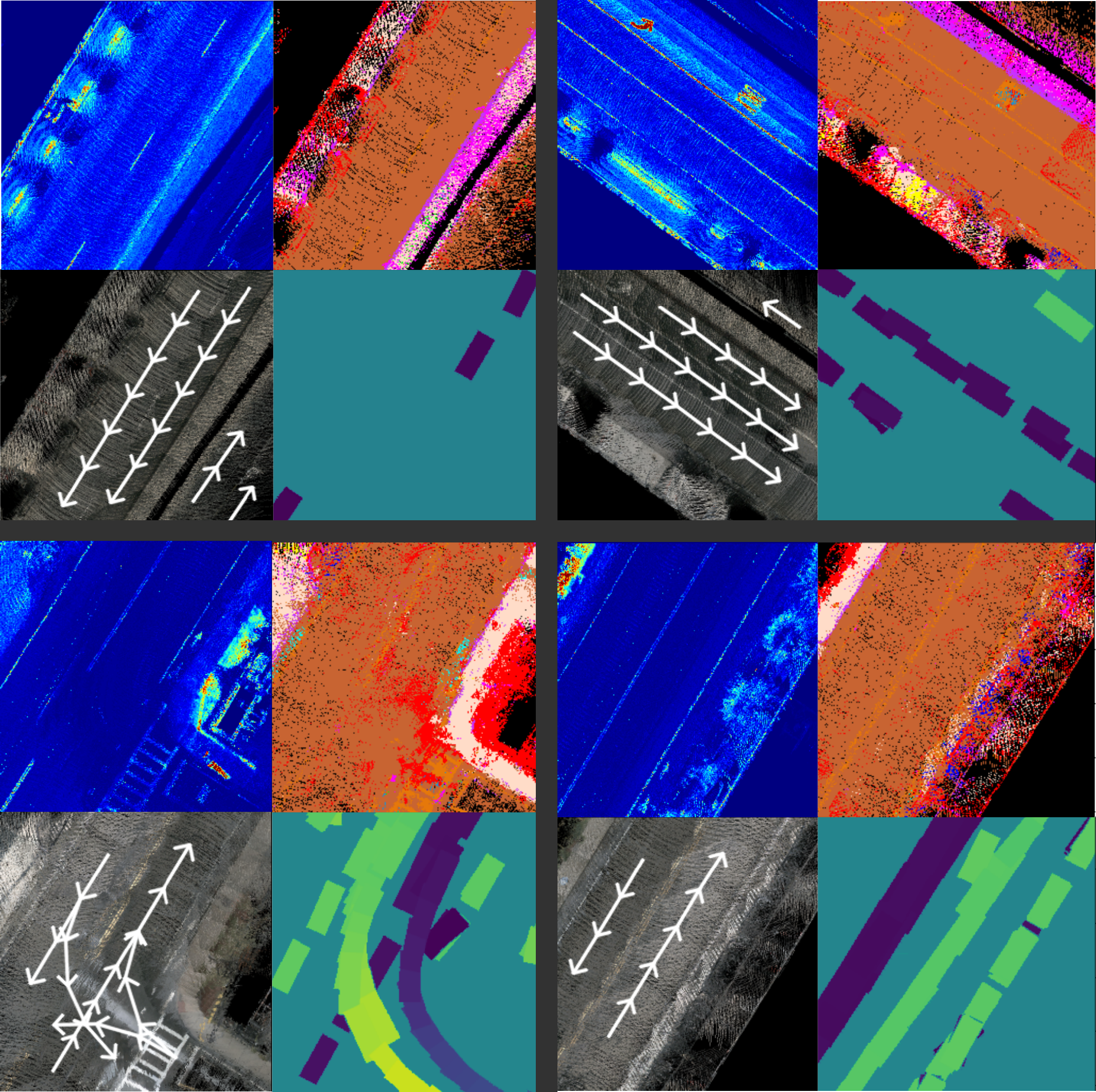}
\caption{Four exemplary sets of BEV modalities from our derived \textit{NuScenesGraph} dataset. The four modalities are: LiDAR intensities (top left), RGB (bottom left), vehicles (bottom right), and semantics (top right). We also visualize the ground truth lane graph superimposed on the RGB image. Note that not for all scenes dense vehicle data is available due to varying traffic density from scene to scene. Color-coding for the semantic classes: Road \raisebox{2pt}{\fcolorbox{black}{road}{\rule{0pt}{1pt}\rule{1pt}{0pt}}}\quad Road Marking \raisebox{2pt}{\fcolorbox{black}{marking}{\rule{0pt}{1pt}\rule{1pt}{0pt}}}\quad Sidewalk \raisebox{2pt}{\fcolorbox{black}{sidewalk}{\rule{0pt}{1pt}\rule{1pt}{0pt}}}\quad Terrain \raisebox{2pt}{\fcolorbox{black}{terrain}{\rule{0pt}{1pt}\rule{1pt}{0pt}}}\quad Curb \raisebox{2pt}{\fcolorbox{black}{curb}{\rule{0pt}{1pt}\rule{1pt}{0pt}}}\quad}
\label{fig:dataset-peek}
\end{figure}

\begin{table}
\centering
\caption{Key data for our derived dataset NuScenes Graph}
\label{tab:dataset-details}
\begin{tabular}{p{1.5cm}p{1cm}p{1cm}p{1.5cm}p{2cm}}
Labeled area  & Vehicles  & Semantic Classes & Lane Segments  \\
 \hline
$\SI{1.446}{km^2}$ & $35,481$ & $14$  & $21,745$  \\
\end{tabular}
\end{table}

\subsection{Lane Graph Estimation}

To infer a given lane graph from the BEV image modalities, we adapt the Graph-RCNN architecture proposed by Yang \etal \cite{yang2018graph}. We modify the originally proposed architecture and add a separate feature extractors for each modality and concatenate the feature maps after the fourth feature extractor convolutional layer. The concatenated feature maps are fed as input into the Region Proposal Network and Relationship Proposal network, respectively, as described by Yang \etal \cite{yang2018graph}. While Graph-RCNNs have previously been primarily used for the task of scene graph estimation, where a relationship graph between detected objects must be deduced from the existence and arrangement of detected objects in a scene, we show that it is possible to use the same framework for the task of lane graph estimation. We also note that in principle a different network architecture from the one we used in our experiments can be plugged into our LaneGraphNet framework. We encode the connection between two lane anchor points (graph nodes) via a \textit{has-connection}-class which is set to $1$ if two anchor points share a graph edge and $0$ if they do not. With this scheme, most graph edges are assigned the label $0$. The anchor points themselves all have the same label \textit{anchor-point}. As such, both the anchor point classification task and the anchor point relationship classification task necessary to produce the lane graph can be treated as a binary classification tasks. Note that the \textit{has-connection}-class includes no notion of direction. The graph direction is determined at a later stage using the output of the sub-model \textit{LaneDirNet}.

\subsection{Adaptive Graph Resampling}

The ground-truth graph is parameterized by a set of anchor points sampled at a fixed but principally arbitrary rate along a  lane. The model should be allowed to place anchor points freely along lanes as long as they are connected along the lane centerline and as long as they align well with the ground truth graph. The model should only be penalized for placing anchor points with an offset from the lane and for connecting nodes incorrectly. To this end, we integrated an adaptive graph parameterization approach that pre-processes the ground truth graph based on the anchor point proposals produced by the ROI-proposal head. Instead of treating anchor points as fixed in the image, we orthogonally project each anchor point proposal onto its nearest lane. These projections serve as the ground-truth anchor points for calculating the loss. Therefore, the position of the anchor point along the lane does not influence the model loss. Formally, we adjust the ground-truth bounding box position according to the following equation:

\begin{equation}
    \mathbf{t}_{proj} = \mathbf{l}_1 + (\mathbf{l}_2 - \mathbf{l}_1) \cdot (\hat{\mathbf{t}} - \mathbf{l}_1)
\end{equation}

where $\hat{\mathbf{t}}$ is the position vector of the predicted bounding box center and $\mathbf{l}_1$ and $\mathbf{l}_2$ denote position vectors for the two anchor points of the closest lane segment in the ground truth graph. Additionally, we reorder the projected proposals and add \textit{has-connection}-labels for pairs of projected points along the lane to obtain the correct connectivity of the graph. Note that even though this adjustment of the ground-truth graph must be done for each set of anchor point proposals during training, the computational overhead for the re-sampling is marginal.

\subsection{Lane Directions}

To be able to predict the direction of lanes, we add an additional network to our LaneGraphNet framework, denoted as \textit{LaneDirNet}. We use the PSPNet architecture \cite{zhao2017pyramid} for this model. The model estimates for each BEV input pixel, in which direction the lane is heading, if it contains a lane. We formulate this estimation task as a classification task, where we bin the direction angles $\theta \in [0^{\circ}, ~360^{\circ}]$ into classes, where each class entails angles $\theta \in [k~20^{\circ}, (k+1)~20^{\circ}], k \in \{0,\ldots,17\}$, resulting in a total of 19 classes. We also include a background class for regions that do not contain any lanes. The model is trained with a per-pixel cross-entropy loss, denoted as $\mathcal{L}_{dir}$:

\begin{equation}
    \mathcal{L}_{dir} = \frac{1}{HW} \sum_{(i,j)} y_k^{(i,j)} \log \hat{p}_k^{(i,j)}
\end{equation}

To obtain the direction of a specific lane section between two anchor points, we convert the class-predictions back to an angle representation, query the predicted lane direction at both lane anchor points, calculate the mean direction angle, and convert it to a vector representation $\Tilde{\bold{d}}$. This vector is then compared to the vector connecting the two lane anchor points $A$ and $B$. For the resulting lane direction, there are only two directions possible: Either $A \xrightarrow{} B$, denoted as $\bold{a^{+}}$ or $B \xrightarrow{} A$, denoted as $\bold{a^{-}}$. We select the direction $\bold{d}$ yielding the largest dot product absolute value:

\begin{equation}
\bold{d} = \argmax_{\bold{a^{+}}, \bold{a^{-}}}(|\Tilde{\bold{d}} \cdot \bold{a^{+}}|, |\Tilde{\bold{d}} \cdot \bold{a^{-}}|)    
\end{equation}

\subsection{Post-Processing}

The lane graph as predicted by our LaneGraphNet typically contains easily detectable false positive connections between lane anchor points. This includes connections between too distant pairs of points or isolated points without any lane segments connected to them. To remove these detections, we apply the following post-processing steps in order:

\begin{itemize}
    \item Anchor points with a detection score below $0.5$ are removed. Lane segments containing a removed anchor point are also removed.
    \item Lane segments with a score below $0.2$ are removed.
    \item Lane segments spanning a distance of more than $1/4$ of the image dimensions are removed.
    \item Anchor points with no lane segment connections are removed.
\end{itemize}

\subsection{Training Details}

The complete Boston-Seaport lane graph spans an area of approximately $7000 \times $7000 meters which translates to $35000 \times 35000$ pixels with our resolution of $\SI{0.2}{m}$ per pixel. Our aim, however, is to extract the lane graph in the vicinity of the vehicle and not the full map. Thus, we extract random graph crops, where each crop has a resolution of $256 \times 256$ pixels, resulting in a crop showing an area of size $\SI{51.2}{m} \times \SI{51.2}{m}$. The map is subsequently split into disjoint training and testing regions based on the map coordinates. The training region  covers an area of ca. $\SI{480}{m} \times \SI{480}{m}$ while the testing region covers ca. $\SI{211}{m} \times \SI{211}{m}$. Note that we only consider areas  that contain LiDAR scans and ignore regions where the recording vehicle did not pass and thus no BEV image data would be available. During training, random dropout is applied to the Semantic, RGB and LiDAR map layer. We also apply random brightness and contrast augmentation on the RGB and LiDAR map layers.

\section{Experimental Results}
\label{sec:experiments}

\subsection{Evaluation Metrics}

To quantify the predictions of our model, we use multiple metrics which quantify the quality of both the position of each lane anchor point as well as the connections between them. We evaluate each metric on a per-sample basis and average over the complete test set. In the following, we detail our evaluation metrics.

\textbf{Average Path Length Similarity (APLS)}: The APLS metric \cite{van2018spacenet} sums the differences in optimal path lengths between nodes in the ground truth graph $G$ and the proposal graph $G^\prime$. Missing paths in the graph are assigned the maximum proportional difference of $1.0$. The APLS metric scales from $0$ (worst) to $1$ (best). Formally, it is defined as

\begin{equation}
    \text{APLS} = 1 - \frac{1}{N_p} \sum_{d(v_1,v_2) < \infty} \min \Big\{ 1, \frac{|d(v_1,v_2) - d(v_1',v_2') |}{d(v_1,v_2)} \Big\}, 
\end{equation}

where $v_i$ and $v_i'$ are nodes in $G$ and $G'$, respectively. $N_p$ denotes the number of nodes in $G$ and $d(\cdot,\cdot)$ is the euclidean distance function.

\textbf{\textbf{Direction Accuracy}}: We determine the directional accuracy of the predicted graph by comparing the angle of the directed graph edge connecting two lane anchor points w.r.t. the global map north with the lane direction angle obtained from the ground truth graph. An edge direction counts as correct if the absolute angle difference $|\Delta| < \frac{\pi}{2}$.

\textbf{Dice-Score (F1)} and \textbf{IoU score}: We render the predicted graph and the ground truth graph as an image with the same resolution as the input map layers and calculate the overlap metrics F1 and IoU scores that are equivalent to the evaluation of a binary segmentation task. Each lane is rendered with a diameter of 1.8 meters (equal to 9 pixels with our selected resolution) which is the standard value for lanes in the NuScenes dataset.

\textbf{Symmetric Chamfer Distance}: Similar to Yang \etal \cite{yang2018graph}, we quantify the distance of the set of predicted lane anchor points $X$ to the set of actual anchor points $Y$ using the symmetric Chamfer distance, defined as:

\begin{equation}
    d(X, Y) = \sum_{x \in X} \min_{y \in Y} ||x-y||^2 + \sum_{y \in Y} \min_{x \in X} ||x-y||^2.
\end{equation}

The optimal Chamfer Distance for a set of graph nodes is $0.0$ and is obtained when all nodes are perfectly overlapping. Note that this metric is the only metric directly measuring the euclidean distance between predicted and ground truth lane anchor points, regardless of their position along the ground truth lane. Therefore, it penalizes placements of anchor points along the ground truth lane when they do not overlap with ground truth anchor points, which is undesirable. Due to the high density of lane anchor points in the ground truth graph, we found this effect to be negligible.

\subsection{Baselines}

Due to the lack of prior work solving the task presented in this work, we cannot compare to previous works. Instead, we create two strong baselines. In our first baseline, B1, we train a lane centerline regression model on lane distance transform images generated from the lane line coordinates. We apply a Gaussian filter to provide additional distance smoothing. We subsequently skeletonize the predicted lane-centerline to obtain a skeleton image where lane centerline pixels have value $1$ and the remaining pixels have value $0$. Finally, we create a lane graph from intersections of skeleton centerlines. In our second baseline, B2, we predict the lane anchor points using the same region proposal network as in the Graph-RCNN backbone but we omit the relationship proposal network. Instead, for each lane anchor point, we connect it with its two closest neighbors. With this point connection heuristic, we build up the lane graph.

\begin{figure}
\centering
\includegraphics[width=\linewidth]{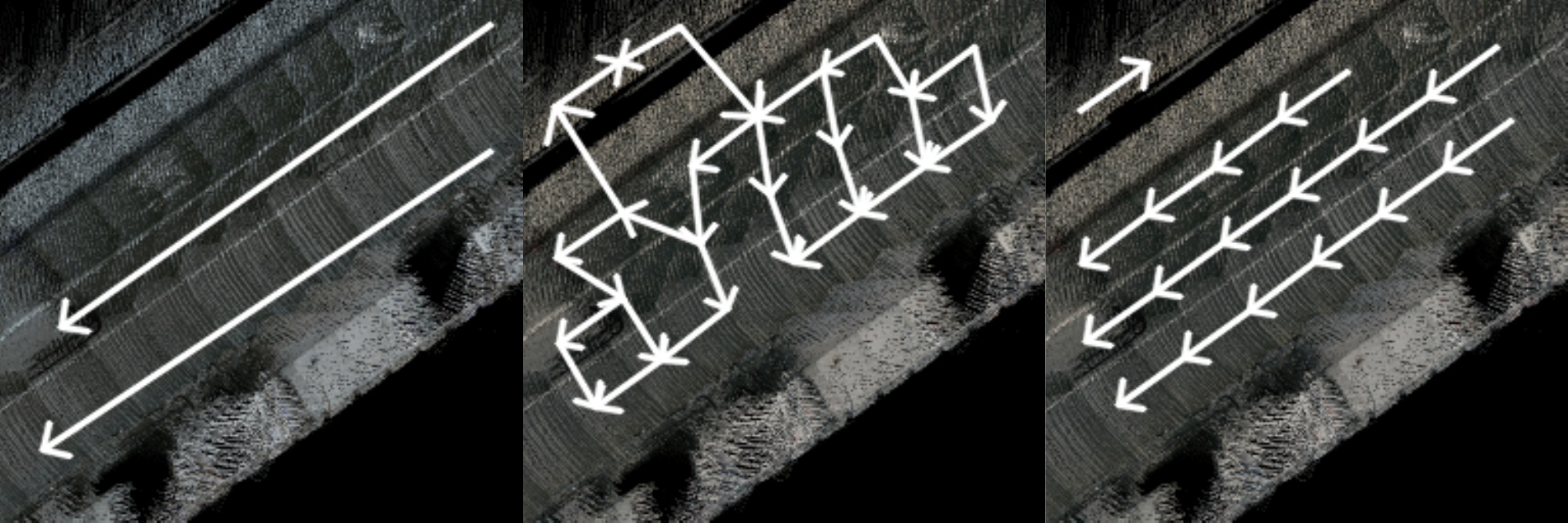}
    \caption{Exemplary predictions of the two baseline models B1 (left) and B2 (center). The ground truth graph is visualized on the right.}
    \label{fig:qualitative-baselines} 
\end{figure}

\begin{table}
\centering
\caption{Quantitative comparison of our approach with baselines}
\label{tab:baselines}
\begin{tabular}{p{4.5cm}p{0.6cm}p{0.6cm}p{0.6cm}}
Model & APLS & F1 & IoU  \\
 \hline
Skeletonized Lane Regression (B1) & $0.033$ & $0.202$ & $0.114$ \\
Anchor Point Nearest Neighbors (B2) & $0.088$ & $0.448$ & $0.298$ \\
 \hline
LaneGraphNet (Ours) & $\mathbf{0.176}$ & $\mathbf{0.574}$ & $\mathbf{0.420}$ \\
\end{tabular}
\end{table}

As demonstrated in Tab. \ref{tab:baselines}, our LaneGraphNet performs better than both baselines in all metrics. While the skeletonized lane regression baseline (B1) performs poorly, the performance of B2 is much closer to the performance of our LaneGraphNet. We observe that lane centerline regression cannot serve as a proxy for explicitly reasoning about the structure of the lane graph. We furthermore hypothesize that in B2, the high precision in placing the lane anchor points allows the nearest-neighbor search to select the correct lane connection with higher accuracy than in B2, albeit it does not reach the accuracy of the LaneGraphNet. Exemplary inference results are illustrated in Fig.~\ref{fig:qualitative-baselines}.  We conclude that a simple heuristic for connecting neighboring anchor points can work reasonably well if they are placed correctly, however, explicitly learning to connect anchor points based on data produces superior results.

\subsection{Ablation Studies}

\begin{table*}
\centering
\caption{Comparison of our LaneGraphNet with its variants}
\label{tab:baselineComp}
\begin{tabular}{p{1cm}|p{0.5cm}p{0.5cm}p{0.5cm}p{2.2cm}|p{0.6cm}p{0.4cm}p{0.8cm}p{0.7cm}p{1.8cm}}
Variant &F1	& IoU & APLS & Chamfer Distance  & 	LiDAR &	RGB	& Semantics & Vehicles	& Input Adapter \\

 \hline
 \hline
V1 & $0.167$ & $0.011$ & $0.041$ &   $24,566.0$ & \checkmark & \xmark & \xmark & \xmark  & \xmark  \\
V2 & $0.458$ & $0.307$ & $0.090$ &   $735.0$ & \checkmark & \checkmark & \xmark & \xmark  & \xmark  \\
V3 & $\mathbf{0.574}$ & $\mathbf{0.420}$ & $0.176$ &   $692.0$ & \checkmark & \checkmark & \xmark & \xmark  & \checkmark \\
V4 & $0.501$ & $0.350$ & $0.109$ &   $643.0$ & \checkmark & \checkmark & \checkmark & \xmark  & \checkmark \\
V5 & $0.491$ & $0.338$ & $0.112$ &   $\mathbf{558.0}$ & \checkmark & \checkmark & \xmark & \checkmark  & \checkmark \\
V6 & $0.549$ & $0.390$ & $\mathbf{0.177}$ &   $1,653.0$ & \checkmark & \checkmark & \checkmark & \checkmark  & \checkmark \\
\end{tabular}
\end{table*}

In order to evaluate the performance of the components of our LaneGraphNet model, we perform ablation studies with different variants of our model. 
We first study the impact of adding the RGB image modality as an additional input modality (V2), compared to using only LiDAR in variant V1. We observe that the RGB data improves the performance of the model by a large margin. As illustrated in Fig.~\ref{fig:dataset-peek}, the RGB modality contains important scne data that is not entailed in LiDAR intensity data alone. For variant V3, we investigate the influence of the modality input adapter on the model performance, which is illustrated in Fig.~\ref{fig:approach}. We observe that the input adapter helps boosting the overall model performance metrics compared to variant V2 without the input adapter. We hypothesize that the concatenation of feature maps as means of modality fusion provides a richer representation for the subsequent downstream task of graph estimation than concatenating the input modalities themselves. We employ the input adapter for all subsequent model variants.

We further add a semantic input modality and the vehicle detection modality for our variants V4 and V5, while in V6, we activate all input modalities. Interestingly, we observe that the inclusion of additional input modalities does not necessarily yield a better performance than only activated RGB and LiDAR modalities (V3). While the smallest Chamfer Distance if obtained with V5, where the vehicles modality is used as an additional input, the variant using all four input modalities (V6) achieves the highest APLS score by a small margin but performs worse than V3 in all other metrics. We hypothesize that the inclusion of the semantic layer adds noisy labels as the predictions of the segmentation model are not perfectly accurate. We further theorize that such noisy labels can lead to worse downstream model performance. We also note that the the density of traffic affects downstream model performance. Additional vehicle data may help a model generalize better to unseen scenes as the position and orientation of vehicles is a useful prior when estimating the position and orientation of lanes, however if the density of traffic varies between scenes, the correlation between the position of vehicles and the lane positions can be subtle and may also worsen the prediction quality. In Fig.~\ref{fig:dataset-peek}, we illustrate the significant difference in traffic density over various scenes.

\subsection{LaneDirNet Performance}

\begin{figure}
\centering
\includegraphics[width=\linewidth]{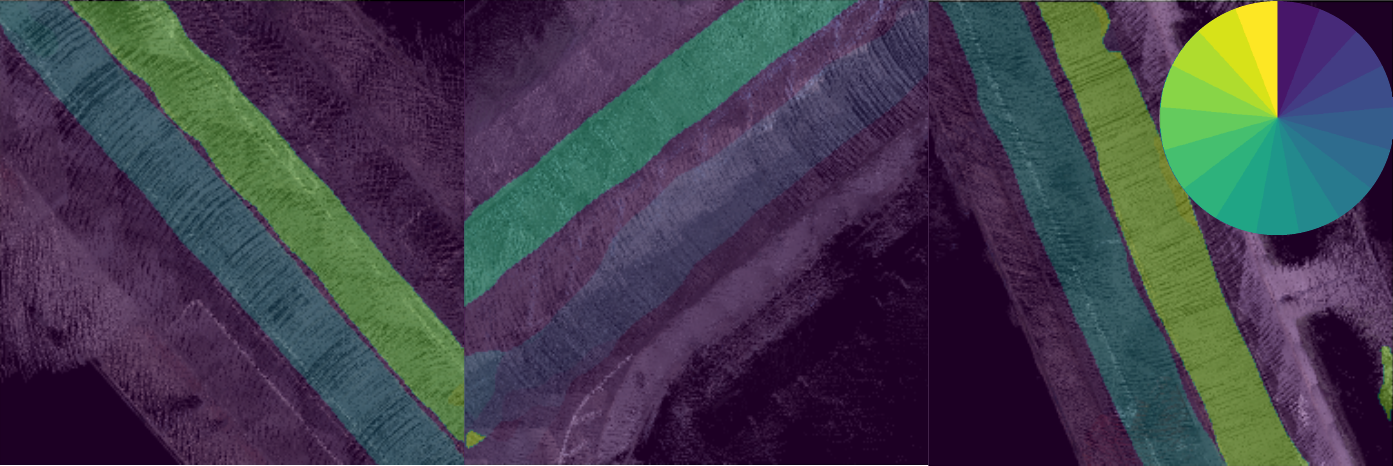}
    \caption{Color-coded exemplary predictions of our LaneDirNet model branch overlaid on BEV RGB images. Directions are assigned to lane segments according to the angle predictions of this model. The class-angle color coding scheme is illustrated by the color circle at the top right.}
    \label{fig:lanedirnet} 
\end{figure}

We further investigate the influence of our LaneDirNet on the direction accuracy of the estimated lane graph. The Graph R-CNN backbone \cite{yang2018graph} used in our approach is principally able to predict the lane direction since we feed the model with a lane segment by providing a source anchor point and a destination anchor point, forming a direction vector. This labeling scheme encodes the graph direction explicitly and should allow the model to infer the direction of each graph edge. We empirically found that this relationship is modelled poorly by the Graph R-CNN backbone present in our LaneGraphNet framework. As described previously, we circumvent this performance bottleneck by explicitly estimating a pixel-wise lane-direction via a pixel-wise classification task using our LaneDirNet model. Figure \ref{fig:lanedirnet} exemplary illustrates predictions of our LaneDirNet where we overlay the color-coded predicted direction on top of RGB images. We quantitatively compare the direction accuracy of the Graph R-CNN backbone with the direction accuracy obtained with input from our LaneDirNet. The Graph-RCNN backbone achieves an overall direction accuracy of $50.8\%$ which is only marginally better than pure chance ($50\%$). Our LaneDirNet, in contrast, achieves an overall direction accuracy of $93.5\%$.

\subsection{Qualitative Results}

\begin{figure*}
\centering
\footnotesize
\setlength{\tabcolsep}{0.0cm}
    \begin{tabular}{P{4.5cm}P{4.5cm}@{\hskip 2mm}P{4.5cm}P{4.5cm}}
    Ground truth & Ours & Ground truth & Ours \\ [0.2cm] \\
           \raisebox{-.5\height}{\includegraphics[width=\linewidth]{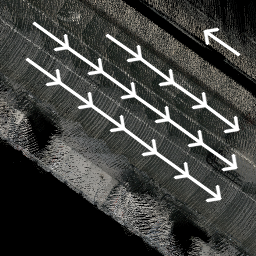}} & \raisebox{-.5\height}{\includegraphics[width=\linewidth]{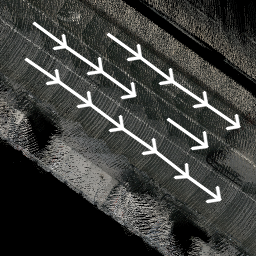}} &   \raisebox{-.5\height}{\includegraphics[width=\linewidth]{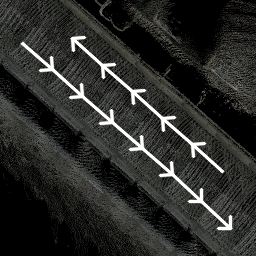}} & \raisebox{-.5\height}{\includegraphics[width=\linewidth]{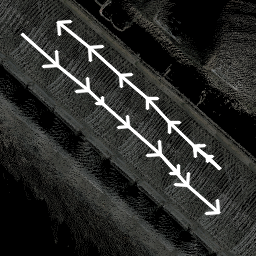}} \\[0.2cm] \\
           \raisebox{-.5\height}{\includegraphics[width=\linewidth]{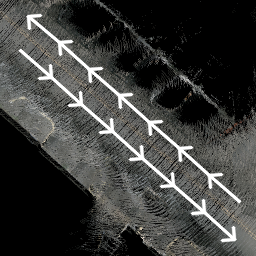}} &  \raisebox{-.5\height}{\includegraphics[width=\linewidth]{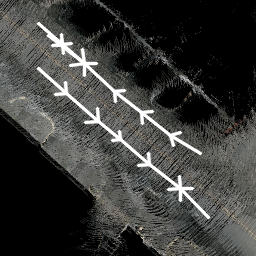}} &  \raisebox{-.5\height}{\includegraphics[width=\linewidth]{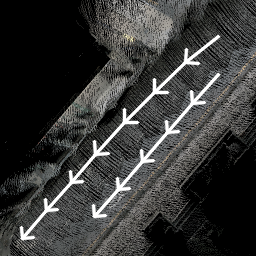}} & \raisebox{-.5\height}{\includegraphics[width=\linewidth]{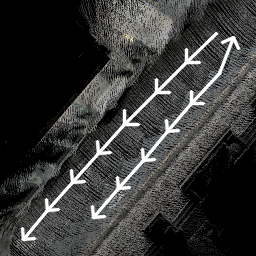}} \\[0.2cm] \\
           \raisebox{-.5\height}{\includegraphics[width=\linewidth]{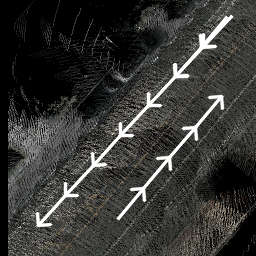}} &  \raisebox{-.5\height}{\includegraphics[width=\linewidth]{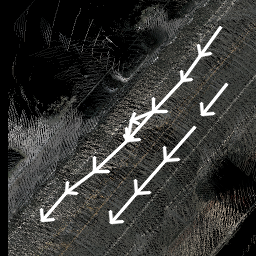}} &  \raisebox{-.5\height}{\includegraphics[width=\linewidth]{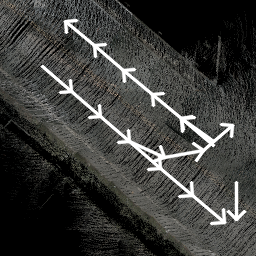}} & \raisebox{-.5\height}{\includegraphics[width=\linewidth]{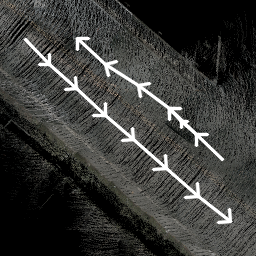}} \\[0.2cm] \\
          \raisebox{-.5\height}{\includegraphics[width=\linewidth]{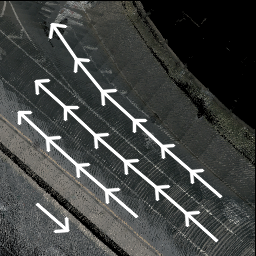}} & \raisebox{-.5\height}{\includegraphics[width=\linewidth]{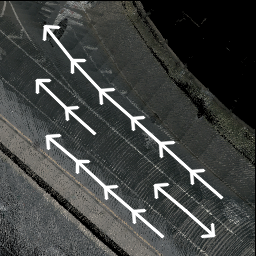}} &  \raisebox{-.5\height}{\includegraphics[width=\linewidth]{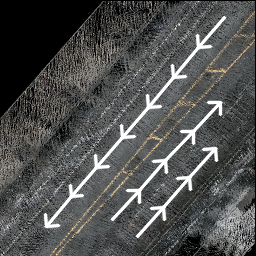}} & \raisebox{-.5\height}{\includegraphics[width=\linewidth]{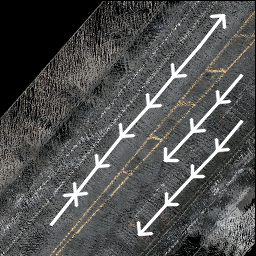}} \\[1cm] \\
          \raisebox{-.5\height}{\includegraphics[width=\linewidth]{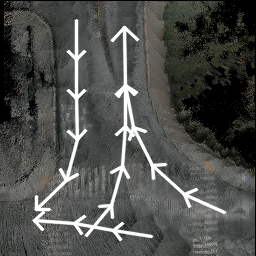}} & \raisebox{-.5\height}{\includegraphics[width=\linewidth]{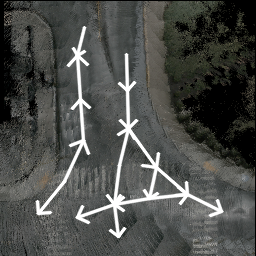}} &  \raisebox{-.5\height}{\includegraphics[width=\linewidth]{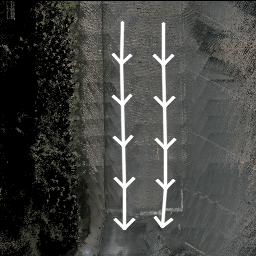}} & \raisebox{-.5\height}{\includegraphics[width=\linewidth]{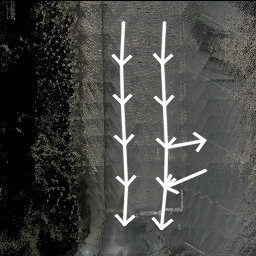}}\\
    \end{tabular}
    \caption{Qualitative results on the test split of the Boston Seaport map. For simplicity, we only visualize the RGB layer of the map and omit the other layers. Images in the two bottom rows illustrate failure cases of our approach.}
    \label{fig:qualitative} 
\end{figure*}

For an overview over qualitative results, refer to Fig.~\ref{fig:qualitative}. In rows 1-3, we illustrate typical results from our LaneGraphNet model after post-processing. We observe that road sections without intersections have high graph quality with correct placement of lane anchor points and mostly correct placement of connections between them, regardless of the number of lanes present in the scene. This applies for straight and curvy road sections. We also observe that some intersection with two or more roads meeting have high visual graph quality.

In rows 4 and 5, we illustrate typical failure cases of our model. We identify multiple interesting failure modes for our model. Firstly, we observe that our LaneDirNet model for predicting the lane direction can break down such that actually opposing lanes are estimated to have the same overall direction (row 4, right scene). We theorize that some BEV input images contain no cues indicating the overall direction of a lane. We also note that our heuristic for filtering false positive lane sections can mistakenly remove lane sections from the graph (row 4, left scene). Moreover, we observe that intersections pose a particularly hard challenge for our model due to the complexity of the underlying lane graph. Thus, some intersections are associated with an ill-formed intersection graph, even though the overall position of lane anchor points is valid. We also note that complex scenes with multiple lanes entering and leaving an intersection remain a challenge. We hypothesize that the limited amount of training data containing such complex intersections prevents the model from generalizing to such intersections. Interestingly, the model also predicts a non-existent intersection arm (bottom row, right scene), where multiple parking lots are located, which visually appear similar to an intersection arm. We also note that our model has learned an accurate lane direction prior as the two false-positive lane segments are associated with the correct directions for right-hand driving.

\section{Conclusion}
\label{sec:conclusion}
In this work, we presented a novel approach for lane graph estimation from multiple input modalities in challenging urban scenes. Our framework is trained to predict both the lane anchor points, the existence of connections between the anchor points, and the direction of the connection for a given scene. We generated a rich urban dataset derived from the popular NuScenes dataset and thoroughly evaluated our approach. We showed qualitatively and quantitatively that our approach performs well in most tested scenes and clearly outperforms strong baselines.



{\small
\bibliographystyle{ieee_fullname}
\bibliography{root}
}

\end{document}